\documentclass[12pt]{article}

\usepackage{silence}
\usepackage[utf8]{inputenc}
\usepackage[T1]{fontenc}
\usepackage{microtype}
\usepackage[margin=25mm]{geometry}
\usepackage{hyperref}
\hypersetup{
    colorlinks=true,
    linkcolor=blue,
    filecolor=magenta,      
    urlcolor=cyan,
    pdftitle={Destruction},
}
\usepackage{titling}
\usepackage[style=authoryear, backend=biber]{biblatex}
\def\BibTeX{{\rm B\kern-.05em{\sc i\kern-.025em b}\kern-.08em T\kern-.1667em\lower.7ex\hbox{E}\kern-.125emX}}
\usepackage{mdframed}
\newmdenv[
  topline=false,
  bottomline=false,
  rightline=false,
  leftline=true,
  linewidth=3pt,
  linecolor=gray,
  leftmargin=50pt,
  rightmargin=50pt,
  innerleftmargin=10pt,
  innerrightmargin=0pt,
  innertopmargin=4pt,
  innerbottommargin=4pt,
  skipabove=\baselineskip,
  skipbelow=\baselineskip,
]{blockquote}

\usepackage{amsmath, amssymb, amsthm}
\usepackage{mathtools}
\usepackage{cancel}
\usepackage{tikz-cd}
\usetikzlibrary{decorations.pathmorphing,fit,calc,shapes.geometric}

\tikzset{
  stoch/.style={->[left]},
  stochto/.style={|->[left]},
  squigglish/.style={decorate, decoration={snake, segment length = 4pt, amplitude=1pt, pre length=1pt, post length=1pt}},
  setellipse/.style={ellipse,draw,inner xsep=1.5ex, inner ysep=-.25ex},
  setbox/.style={draw,inner xsep=1.5ex, inner ysep=-.25ex, rounded corners},
  commutative diagrams/diagrams={arrows={shorten >=-.5ex,shorten <=-.5ex}},
}

\usepackage[capitalise, noabbrev]{cleveref}

\title{Destruction is a General Strategy to Learn Generation;\\Diffusion's Strength is to Take it Seriously;\\Exploration is the Future}
\author{Pierre-André Noël \\ \small ServiceNow AI Research}
\date{Published April 27th, 2026 as an ICLR blogpost {\small\url{https://iclr-blogposts.github.io/2026/blog/2026/destruction/}}}

\begin{document}
\maketitle
\begin{abstract}
I present diffusion models as part of a family of machine learning techniques that withhold information from a model's input and train it to guess the withheld information. I argue that diffusion's destroying approach to withholding is more flexible than typical hand-crafted information withholding techniques, providing a rich training playground that could be advantageous in some settings, notably data-scarce ones. I then address subtle issues that may arise when porting reinforcement learning techniques to the diffusion context, and wonder how such exploration problems could be addressed in more diffusion-native ways. I do not have definitive answers, but I do point my fingers in directions I deem interesting. A tutorial follows this thesis, expanding on the destroy-then-generate perspective. A novel kind of probabilistic graphical models is introduced to facilitate the tutorial's exposition.
\end{abstract}
\tableofcontents

\begin{blockquote}
Pourquoi faire simple quand on peut faire compliqué?\smallskip

\hfill-- French locution, usually ironic.
\end{blockquote}

This blogpost is composed of two main sections, tied together by their unusual information-theoretic viewpoint.
\emph{The Thesis} (Section \ref{section:thesis}) is an opinion/perspective/speculative piece on diffusion models.
\emph{The Tutorial} (Section \ref{section:tutorial}) is a diagrammatic presentation of diffusion models.

\textbf{For diffusion experts.}
\emph{The Thesis} is meant to stand alone, and it is where you will find the bulk of this blogpost's novel ideas.
Nonetheless, I encourage you to at least browse \emph{The Tutorial}: it takes an unusual perspective that may interest you, and it introduces a new kind of probabilistic graphical model -- which I propose to call \emph{generative commutative diagrams} -- that personally helped me wrap my head around numerous subtleties.

\textbf{For non-experts.}
I encourage you to read through \emph{The Thesis} in a cursory first-pass, just to absorb its general sense, then immediately proceed with \emph{The Tutorial}.
This should prepare you to circle back to \emph{The Thesis}, this time paying more attention to passages that give you pause.

Finally, I wish to remind all readers that \textbf{this blogpost contains speculations}: I've taken special care toward clearly marking them as such, and I hope that this callout helps placing you in the right mindset.

\section{The Thesis\label{section:thesis}}
\subsection{Learning by Destroying}

\textcite{yang2023diffusion} defined generative diffusion models as ``a family of probabilistic generative models that progressively destruct data by injecting noise, then learn to reverse this process for sample generation.''
I think that it is a good definition, except for the ``by injecting noise'' part.
Indeed, there is a paper that is literally called \emph{Cold Diffusion: Inverting Arbitrary Image Transforms \textbf{Without Noise}} \autocite[emphasis mine]{bansal2023cold}, and calling ``noise'' the action of the \texttt{[MASK]} token in Mask Diffusion Language Models~\autocite[MDLMs]{sahoo2024simple} is certainly debatable.
For the sake of this blogpost, I will define generative diffusion models as follows.

\begin{enumerate}
\item \textbf{Data distribution.} We have a training dataset sampled from a data distribution, and our goal is to generate new samples from the same data distribution. The only allowed ``control mechanism'' is \emph{conditioning}.\footnote{Concretely, the condition can be specified as a predicate: a function accepting a data sample and returning either ``true'' or ``false''. When such a condition is provided, the model must only return samples for which the predicate returns ``true'', otherwise keeping the data distribution unaltered. Notice that this filtering could be achieved by wrapping the model in a loop, and returning the first samples that passes the predicate check. More generally, the same argument holds for stochastic (leaky) filters.}
\item \textbf{Destroying process.} We also have access to a procedure that can gradually destroy the information in data samples. This procedure \textbf{may} involve randomness. Like the training dataset, this procedure is fully specified \emph{a priori}, before any training takes place.
\item \textbf{Generative process.} We train a generative process so that the samples it generates approximate the sought data distribution. This training leverages both the training dataset, which specifies \emph{what} the generative process should ultimately generate, and the destroying process, which specifies \emph{how} to chunk generation into manageable pieces.\footnote{The loss function being minimized is typically an expectation of an expectation of an expectation, with those expectations respectively taken over the training data sample, the ``level of destruction'' to be applied on this training data sample, and the exact stochastic realization of this destruction (if applicable).}
\end{enumerate}

Using this definition, diffusion models can be seen as a special case of the tried-and-true machine learning technique consisting of withholding some information from a model's input, then training the model to generate the withheld information.
\begin{itemize}
\item In supervised learning, the withheld (then generated) information is called ``label''.
\item In an autoregressive language model, it is the next token.
\item For BERT-like language models, it is the \texttt{[MASK]}ed tokens.
\item Some autoencoders rely on a representational bottleneck to withhold information.
\item Contrastive autoencoders withhold whether pairs of samples are somehow related (positive pairs) or not (negative pairs).
\item Denoising autoencoders add ``noise'' to samples, which destroys some of their informational content (\emph{i.e.}, withholding it).
\item Diffusion models use the destroying process to withhold information.\footnote{This typically involves some form of randomness, but here we do not make it a requirement.}
\end{itemize}

You may notice a connection between the last two: that connection has already been made long ago~\autocite{kingma2021variational}.
In fact, such connections have been made between diffusion and other points in that list \autocite{zheng2025masked,fathi2025unifying}, and I think that, given enough coffee, we could get them all covered.
\emph{Should we}?
That's a good question, and the answer may depend in part on how data-starved we are.

\subsection{Diffusion's Advantages}

In \emph{Diffusion Language Models are Super Data Learners}, \textcite{ni2025diffusion} observe that

\begin{blockquote}
when unique data is limited, diffusion language models (DLMs) consistently surpass autoregressive (AR) models by training for more epochs. The crossover shifts later with more or higher-quality data, earlier with larger models, and persists across dense and sparse architectures. We attribute the gains to three compounding factors: (1) any-order modeling, (2) super-dense compute from iterative bidirectional denoising, and (3) built-in Monte Carlo augmentation [\ldots]
\end{blockquote}

Let's speculate on the situation.
We have devised simple (non-diffusion) ways to withhold information from our models, breaking the data into chunks that ``make sense'' to us, emphasizing aspects that must obviously be learned sooner or later.
And when we train models to generate that withheld information, these simple approaches may turn out to be locally optimal for the sake of training as quickly as possible on humongous datasets.
But when the fresh data gets scarce, when you have to train from the same sample for the $N^{\text{th}}$ time, perhaps more could be learned by considering a different viewpoint -- like a diffusion model learning to predict tokens out-of-order.
To be clear, ultimately, it could be that the diffusion model will perform better during inference when it is used to predict tokens in order, like an autoregressive model would simply do~\autocite{kim2025train}.
But because it was \emph{challenged} during training on out-of-order tasks, it may eventually manage to pick up some tricks that will forever evade the autoregressively-trained model.
Bigger models have more capacity to latch on such tricks, so their crossover may come earlier.
\textbf{If these speculations are on the right path, then we can predict this phenomenon to be general}: among different ways to learn-by-destroying, diffusion-like approaches to destruction should outperform clean/systematic ones in data-starved regimes.\footnote{Given enough training epochs and model capacity.}\footnote{Hence the French locution opening this blogpost.}

Depending on your perspective, this may align with \emph{The Bitter Lesson}~\autocite{sutton2019bitter}:

\begin{blockquote}
general methods that leverage computation are ultimately the most effective, and by a large margin. [\ldots] Seeking an improvement that makes a difference in the shorter term, researchers seek to leverage their human knowledge of the domain, but the only thing that matters in the long run is the leveraging of computation.  
\end{blockquote}

Left-to-right training may have been yet another human bias, a temporary fad, and in the long run perhaps we'll just train in all-the-orders, leveraging much more compute.
But there is a sweeter perspective on that bitter lesson: the design space for novel diffusion models is huge.
To this day, very few destroying processes have been considered.
For example, \texttt{[MASK]} is a great candidate for the next human fad/bias to be replaced by something that better leverages compute.
Moreover, while diffusion may beat our vanilla information-withholding approaches in the bounded data regime, it hasn't been shown to be the best way to proceed: diffusion itself may be a weird human fad.

\subsection{Reconciling Exploration}

This takes me to the last point of this blogpost's thesis: if all incarnations of the tried-and-true ``information withholding'' machine learning technique can (and perhaps should) be related to diffusion, what other techniques are left?
And could we improve upon diffusion by learning from them?
The first answers that come to my mind are ``anything that involves exploration,''\footnote{Here by ``exploration'' I roughly mean ``trying things'': whenever ``frozen accidents'' could induce path dependences in the details of the model's training history.} and ``yes, probably.''

The archetypal technique involving exploration is Reinforcement Learning (RL).
Whereas pure generative models solely learn from a training dataset -- striving to generate new samples from the exact same data distribution -- RL models strive for a different, ``better'' distribution.
What is meant by ``better'' is usually (but not always) specified through a reward function: a function that assigns value (reward) to each possible data sample.
We typically wish to maximize the expected reward, though there may be some requirement to not meander too far away from the original data distribution.

GRPO~\autocite{shao2024deepseekmath} is a quite successful RL technique, and it has recently been ported to diffusion~\autocite{gong2025diffucoder}.
Such an adaptation is not trivial: due to its autoregressive origin, GRPO takes for granted easy access to likelihoods for different sequences, but such likelihoods are harder to get in diffusion models.
This issue has been overcome before in the context of perplexity~\autocite{sahoo2024simple}: take expectations over different levels of destruction and over different destruction realization, which translates to different decoding orders.
\textcite{gong2025diffucoder} leverage this technique,\footnote{With the twist that when a decoding order is used in the estimation of this expectation, the ``complement'' of the mask associated to that decoding order is always also considered.
This practice reduces the variance, allowing them to take expectations over a single such pair of orders.}
and their results are great!

Yet I suspect that this expectation-over-decoding-order strategy is inherently off-policy.
Indeed, even for purely random decoding orders,\footnote{To be clear, I definitely believe that it would be a good idea to learn a policy for the decoding order. I'm here assuming a non-learnable random order to make my point stronger.} I claim that the specific order faced by the model while generating a trace should be accounted for in the reward assignment for that trace.
My intuition goes as follows: if a successful reasoning trace summarizes some premises from the context, expands some methodic steps, then reaches some conclusion, what are we teaching the model by rewarding it to predict the conclusion first, without the reasoning steps that lead to it?\footnote{My personal guess, derived from intuition alone: at best we're teaching it to skip steps and/or ignore reasoning, at worst we're teaching it to hallucinate, make stuff up, justify \emph{a posteriori} and/or otherwise deceive.}

You may now think ``Who are you to say how the model should or shouldn't reason! Remember The Bitter Lesson!''
Fair enough, but here's my point: in the non-RL diffusion case, this expectation-over-decoding-order approach was grounded in sound theory, but we didn't do our homework before porting it to RL.
We can justify it by its good empirical results, but we lost our theoretical grounding.
\emph{A priori}, the only path we may reward on-policy for a given generated sample is the one that was followed by the model while generating that sample.\footnote{In the end, the extent to which this ``off-policy-ordering'' matters (if at all) is a question to be resolved empirically. This implies biting the bullet and doing GRPO with a forward pass for each token in a trace (for both the original model and RLed one, to get ratios), and compare how test performances vary when using the same order as when the trace was generated, versus a new random one. Different downstream tasks may behave differently, and adding structure to the random decoding order (\emph{e.g.}, block decoding) may also affect the outcome.}

Ok, then how did we get that theoretical grounding in the non-RL case?
Limiting ourselves to my strict definition at the beginning of this blogpost, we consider a specific data distribution and noising process, and there thus exists a single, ideal, typically untractable\footnote{This is why we have to train a neural network: we're learning a tractable function that approximates the untractable ideal one.} probability distribution for partially-destroyed data samples at different levels of destruction.
Within a given modeling paradigm, all concrete diffusion model implementations seek to approximate some function of that same ideal distribution: models with different weight initialization (or even different neural network architectures!) all strive to approximate the same ``correct'' answer.
\textbf{This is a very convenient property}: in a diffusion model (as per my definition), the function to be learned is not a moving target.\footnote{For example, some models learn the score function (also know as ``informant''), which approximates the gradient of the log of the aforementioned ideal probability distribution. Other models approximate the expected Gaussian noise that was added to a noisy sample, or a probability distribution over the original clean sample: these are also deterministic (but untractable) functions of the ideal probability distribution.}

Can we approach the RL problem with a diffusion model that satisfies this strict definition?
Yes!
By reframing it as conditioning~\autocite{yuan2023reward,levine2018reinforcement,ziegler2019fine},\footnote{From this viewpoint, \emph{reward tilting} amounts to softly conditioning on high reward via an exponential weighting.} which is the sole allowed control mechanism as per my definition.
For now, concrete work on that front is still in an early stage, and competing approaches leveraging techniques ported from the autoregressive case (\emph{e.g.}, GRPO) have an head start.
Moreover, there is no guarantee that approaching RL by sticking to my strict definition of diffusion has real advantages in the long run.
Nonetheless, I think that the fact we \emph{can} suffices to justify additional exploratory efforts.

More generally, notice how reframing as a conditioning problem removed ``exploration'' from the picture: we're not \emph{searching} for high-reward samples, we're just \emph{filtering out} from the original data distribution the samples that have low reward.\footnote{There is no room for ``frozen accidents'' nor any kind of path dependence. Together, the data distribution and the reward function fully specify the ideal function which the model should strive to approximate.}
In practice, because we don't have infinite resources, actual implementations still have to explore to find where it is worth it to learn the function.
We have the guaranteed existence of a non-moving target function, but we have to find which parts of that target function are worth learning well.\footnote{Recall that ``standard'' RL models are often constrained to not meander too far away from the original data distribution. Concretely, this is usually implemented by comparing the predictions of a frozen, ``old'' model with those of its RLed counterpart. In the conditioned diffusion approach to RL, the non-moving target function plays the role of that old model.}

The astute readers may notice that we've been infringing on a second taboo of my strict definition: the destroying process is \emph{given} (not learned!) before any training takes place, and it specifies what/how the model should learn.
Again, there are workarounds: we may tweak how we sample the destruction process while maintaining the same nice, non-moving target.\footnote{For example, we may counterbalance frequency biases by inversely weighting the losses.}
But is there a point where we could gain something by giving up that nice, non-moving target?

Of course!
For a start, we effectively move that target every time a researcher comes up with a different destroying process, and there is no good reason to believe that this human-in-the-loop algorithm has already found the optimum.
We should \emph{learn} the destroying process!
The data should speak for itself!
The compute should be leveraged!
Umh, right\ldots but how?

Let me start building toward an eventual solution path by formulating a version of this learn-the-destroying-process problem more explicitly.

\begin{enumerate}
\item \textbf{First distribution.} We have a data distribution from which we can get a training dataset.
\item \textbf{Second distribution.} We have a destroying process whose specifics depends on parameters to be determined. However, whatever these parameters are, we already know what the probability distribution for a ``fully destroyed'' data distribution should look like.\footnote{For example, a Gaussian.}
\item \textbf{Cost function.} For given parameters of the destroying process, we train a diffusion model to generate from the data distribution. We are given a function that associates a cost to these parameters: the cost may depend on the ultimate performances of the model when transporting from the fully-destroyed distribution to the data distribution, but also on the model's size and the resources it consumed at training and/or inference. We seek to minimize the expectation of this cost.
\end{enumerate}

In the special case where the ``destroying process'' is constrained to be a bijection (so it does not destroy information, it is just a reversible mangling\footnote{Actually, it can be understood as an optimal compression for communicating samples from the first distribution through a channel that has been optimized for the second distribution.}) and the cost function may only depend ``mathematically'' on the bijection's parameters (\emph{i.e.}, there are no explicit dependencies on model performances and other implementation details), this is known as an Optimal Transport (OT) problem, which are traditionally introduced with piles of dirt.

Imagine a pile of dirt whose height profile represents a probability distribution (higher probability means more dirt piled up there).
We want to move the dirt around so that it represents a second distribution instead.
There are many ways to do so, many exact plans for where to pick each shovelful of dirt and where to toss them.
We can assign a cost\footnote{That is, a negative reward.} to each such plan, and we must explore the space of plans to find the cheapest.\footnote{If you have ever found yourself sitting in a conference room while the speaker said ``\ldots Wasserstein metric, also known as earth mover's distance\ldots'', then the talk was likely about OT.}

Now the requirement that no information is destroyed amounts to demanding a plan that is perfectly specified, perfectly invertible, and perfectly executed on, down to the very grain of sand.
In the more general case where the destroying process may actually destroy information, we have an \emph{entropy-regularized} OT problem~\autocite{debortoli2021diffusion}.\footnote{If the same conference speaker mentioned ``Schrödinger bridges'', then the talk was likely about entropy-regularized OT.}
This regularization favors plans that can account for some ``splatting'' when a shovelful is tossed: tighter splats incur higher penalty.\footnote{The word ``splat'' should here summon the image of uncontrolled/messy dirt tumbling. There is no way to perfectly ``unsplat'' a shovelful: information is destroyed.}

Back to the point, our learn-the-destroying-process problem can thus be connected to entropy-regularized OT if we allow the cost function to depend on implementation-specific metrics during training and/or inference (\emph{e.g.}, model performances and resource consumption).
Is there anything practical to gain by framing the search for the best destroying process in such terms?
Well, I don't know.
Again, I think that the fact we \emph{can} suffices to justify some exploratory efforts.

So destruction is a general strategy for learning to generate information about a data distribution, diffusion's \cancel{messiness} richness may advantageously leverage destruction in some contexts (including data-starved ones), exploring how to optimize a function doesn't neatly fit this pictures, but there are workarounds reconciliating diffusion and exploration, opening a multitude of avenues for future work.
There are no promises, but I think that we should give it a try.

\section{The Tutorial\label{section:tutorial}}
\subsection{Destroying and Generating}

This section provides a diagrammatic introduction to ``destroying'' and ``generating'' information.

As is tradition in information theory, we consider two characters, Alice and Bob.
Alice communicates one of 3 messages to Bob through a communication channel.
An arrow like this $\rightarrow$ indicates the the channel itself, here \texttt{Identity}, while arrows like that $\mapsto$ indicate the specific action of that channel, \emph{i.e.} what message does Bob receive for each message that Alice may utter.

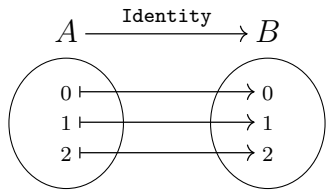
\begin{equation*}
\begin{tikzcd}[
    row sep=tiny, 
    column sep=huge,
    execute at end picture={
        \node[setellipse, fit={(\tikzcdmatrixname-2-1) (\tikzcdmatrixname-4-1)}]{};
        \node[setellipse, fit={(\tikzcdmatrixname-2-2) (\tikzcdmatrixname-4-2)}]{};
    }
]
A \arrow[r, "\texttt{Identity}"] & B \\[.5ex]
{\scriptstyle0} \arrow[r, mapsto] & {\scriptstyle0} \\[-1.5ex]
{\scriptstyle1} \arrow[r, mapsto] & {\scriptstyle1} \\[-1.5ex]
{\scriptstyle2} \arrow[r, mapsto] & {\scriptstyle2}
\end{tikzcd}
\end{equation*}%

In the above situation, Bob gets exactly the message that Alice tried to convey.
The information communicated by Alice is thus preserved (\emph{i.e.} neither created nor destroyed) by the \texttt{Identity} communication channel: it is an important (albeit boring) channel.

Ok, now let's try a different channel, which I'll call \texttt{Cypher}.

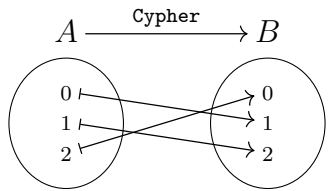
\begin{equation*}
\begin{tikzcd}[
    row sep=tiny, 
    column sep=huge,
    execute at end picture={
        \node[setellipse, fit={(\tikzcdmatrixname-2-1) (\tikzcdmatrixname-4-1)}]{};
        \node[setellipse, fit={(\tikzcdmatrixname-2-2) (\tikzcdmatrixname-4-2)}]{};
    }
]
A \arrow[r, "\texttt{Cypher}"] & B \\[.5ex]
{\scriptstyle0} \arrow[dr, mapsto] & {\scriptstyle0} \\[-1.5ex]
{\scriptstyle1} \arrow[dr, mapsto] & {\scriptstyle1} \\[-1.5ex]
{\scriptstyle2} \arrow[uur, mapsto] & {\scriptstyle2}
\end{tikzcd}
\end{equation*}%

Here Bob gets a different number than the one Alice intended to send.
Does this mean that \texttt{Cypher} destroys information?
Well, it would if the channel could only be used once before being discarded\ldots

But for the sake of this blogpost, let's instead consider the case where Bob is allowed as many training examples as he needs.
During that training, Bob gets Alice's message through both the trusty \texttt{Identity} channel and the to-be-figured-out \texttt{Cypher} channel.
Using pairs of the form $(a,b)$ (with $a \in A$ and $b \in B$), Bob's observations may look like

$$
\{ (0,1), (2,0), (0,1), (0,1), (1,2), (0,1), (2,0), (0,1), (2,0), (0,1), (0,1), (0,1), \cdots \}.
$$

There are many kinds of things that Bob could learn from such data.
First, he may hypothesize that Alice is more likely to give the message $0$ than she is to say $1$ or $2$.
In doing so, Bob would be building a mental model of an Imaginary-Alice $A'$, \emph{i.e.} trying to learn the probability distribution $\textup{P}^\theta(A')$ so that it matches the real $\textup{P}(A)$.
And ultimately, this is exactly what generative modeling is about: learn how to sample from a $\textup{P}^\theta(A')$ that approximates as best as we can the data distribution $\textup{P}(A)$.

Doing so may be realistic for $3$ messages, but what if Alice had more range, say, $10^{678000}$ possible unique messages?\footnote{GPT OSS 120B's vocabulary size powered to its context length.}
This direct approach won't scale, which is why the next section will consider a divide-and-conquer approach to chunk large problems into more amenable ones.
For now, it suffices to say that Bob is just a step in a long chain of messages, followed by Carol, David, \emph{etc.}
If each step is easier to model/learn than the previous one, we'll be making progress toward our ultimate goal.

So, Bob should focus on figuring out the \texttt{Cypher} channel, \emph{i.e.} learn a probability distribution $\textup{P}^{\theta}(A \vert B)$
that approximates $\textup{P}(A \vert B)$.

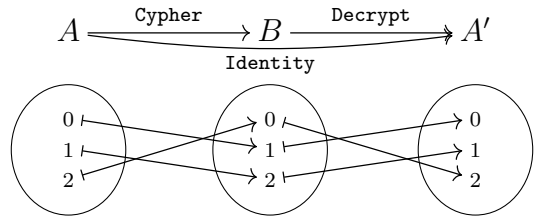
\begin{equation*}
\begin{tikzcd}[
    row sep=tiny, 
    column sep=huge,
    execute at end picture={
        \node[setellipse, fit={(\tikzcdmatrixname-2-1) (\tikzcdmatrixname-4-1)}]{};
        \node[setellipse, fit={(\tikzcdmatrixname-2-2) (\tikzcdmatrixname-4-2)}]{};
        \node[setellipse, fit={(\tikzcdmatrixname-2-3) (\tikzcdmatrixname-4-3)}]{};
    }
]
A \arrow[r, "\texttt{Cypher}"] \arrow[rr, "\texttt{Identity}"', bend right=7] & B \arrow[r, "\texttt{Decrypt}"] & A' \\[2.5ex] 
{\scriptstyle0} \arrow[dr, mapsto] 
& {\scriptstyle0} \arrow[ddr, mapsto] & {\scriptstyle0} \\[-1.5ex]
{\scriptstyle1} \arrow[dr, mapsto] 
& {\scriptstyle1} \arrow[ur, mapsto] & {\scriptstyle1} \\[-1.5ex]
{\scriptstyle2} \arrow[uur, mapsto] 
& {\scriptstyle2} \arrow[ur, mapsto] & {\scriptstyle2}
\end{tikzcd}
\end{equation*}%

There exists a \texttt{Decrypt} channel that, applied on the output of the \texttt{Cypher}, gives the \texttt{Identity}.
If Bob is a good Bayesian, he will never be \emph{absolutely sure} that he figured it out, but whatever his priors were\footnote{Well, within reason, assuming that Bob has a minimal pragmatism and/or understanding of the world\ldots} for the probability distribution from which the \texttt{Cypher} channel was sampled, there is a point at which \texttt{Decrypt} will become his leading hypothesis as to what $\textup{P}^\theta(A' \vert B)$ should be.
And from that point on, his confidence in that hypothesis will keep increasing as more data is gathered.

The existence of a \texttt{Decrypt} that reverses the action of \texttt{Cypher} proves that \texttt{Cypher} does not destroy information.
And because we could do the opposite, \emph{i.e.} reverse the action of \texttt{Decrypt} by applying \texttt{Cypher}, we know that \texttt{Cypher} does not generate information either.
Just like \texttt{Identity}, \texttt{Cypher} preserves information.
What does this mean?
It means that \texttt{Cypher} is a useless channel for our divide-and-conquer aims: learning a $\textup{P}^\theta(B')$ that approaches $\textup{P}(B)$ is as hard as learning a $\textup{P}^\theta(A')$ that approaches $\textup{P}^\theta(A)$.

Ok, maybe what we need is a noisy channel?

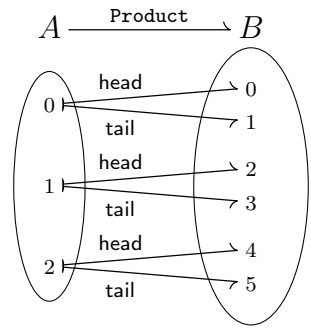
\begin{equation*}
\begin{tikzcd}[
    row sep=tiny, 
    column sep=huge,
    execute at end picture={
        \node[setellipse, fit={(\tikzcdmatrixname-3-1) (\tikzcdmatrixname-11-1)}, inner xsep=2pt, inner ysep=-7pt]{};
        \node[setellipse, fit={(\tikzcdmatrixname-2-2) (\tikzcdmatrixname-12-2)}, inner ysep=-7pt]{};
    }
]
A \arrow[r, "\texttt{Product}", harpoon] & B \\[.5ex]
& {\scriptstyle0} \\[-2.5ex]
{\scriptstyle0} \arrow[ur, mapsto, harpoon, "\textsf{head}"] \arrow[dr, mapsto, harpoon', "\textsf{tail\ }"'] & \\[-2.5ex]
\phantom{\scriptstyle0} & {\scriptstyle1} \\[-2.5ex]
\phantom{\scriptstyle0} & \\[-1.5ex]
\phantom{\scriptstyle0} & {\scriptstyle2} \\[-2.5ex]
{\scriptstyle1} \arrow[ur, mapsto, harpoon, "\textsf{head}"] \arrow[dr, mapsto, harpoon', "\textsf{tail\ }"'] & \\[-2.5ex]
\phantom{\scriptstyle0} & {\scriptstyle3} \\[-2.5ex]
\phantom{\scriptstyle0} & \\[-1.5ex]
\phantom{\scriptstyle0} & {\scriptstyle4} \\[-2.5ex]
{\scriptstyle2} \arrow[ur, mapsto, harpoon, "\textsf{head}"] \arrow[dr, mapsto, harpoon', "\textsf{tail\ }"'] & \\[-2.5ex]
& {\scriptstyle5}
\end{tikzcd}
\end{equation*}%

Here the \texttt{Product}\footnote{This name is a reference to category theory; please see the very last section for why.} channel flips a coin, and this affects what message Bob receives.
Notice that I represented \texttt{Product} using an harpoon\footnote{Whether the harpoon's ``barb'' points up or down is purely aesthetic.} $\rightharpoonup$ instead of an arrow $\rightarrow$, because I reserve arrows for deterministic functions (and \texttt{Product} isn't one).
Does \texttt{Product} destroy information?

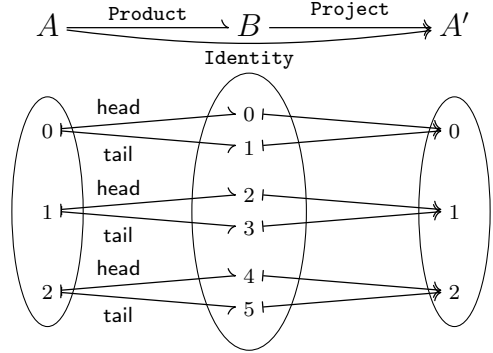
\begin{equation*}
\begin{tikzcd}[
    row sep=tiny, 
    column sep=huge,
    execute at end picture={
        \node[setellipse, fit={(\tikzcdmatrixname-3-1) (\tikzcdmatrixname-11-1)}, inner xsep=2pt, inner ysep=-7pt]{};
        \node[setellipse, fit={(\tikzcdmatrixname-2-2) (\tikzcdmatrixname-12-2)}, inner ysep=-7pt]{};
        \node[setellipse, fit={(\tikzcdmatrixname-3-3) (\tikzcdmatrixname-11-3)}, inner xsep=2pt, inner ysep=-7pt]{};
    }
]
A \arrow[r, "\texttt{Product}", harpoon] \arrow[rr, "\texttt{Identity}"', bend right=7] & B \arrow[r, "\texttt{Project}"] & A' \\[2.5ex]
& {\scriptstyle0} \arrow[dr, mapsto] & \\[-2.5ex]
{\scriptstyle0} \arrow[ur, mapsto, harpoon, "\textsf{head}"] \arrow[dr, mapsto, harpoon', "\textsf{tail\ }"'] & & {\scriptstyle0} \\[-2.5ex]
\phantom{\scriptstyle0} & {\scriptstyle1} \arrow[ur, mapsto] \\[-2.5ex]
\phantom{\scriptstyle0} & \\[-1.5ex]
\phantom{\scriptstyle0} & {\scriptstyle2} \arrow[dr, mapsto] \\[-2.5ex]
{\scriptstyle1} \arrow[ur, mapsto, harpoon, "\textsf{head}"] \arrow[dr, mapsto, harpoon', "\textsf{tail\ }"'] & & {\scriptstyle1} \\[-2.5ex]
\phantom{\scriptstyle0} & {\scriptstyle3} \arrow[ur, mapsto] \\[-2.5ex]
\phantom{\scriptstyle0} & \\[-1.5ex]
\phantom{\scriptstyle0} & {\scriptstyle4} \arrow[dr, mapsto] \\[-2.5ex]
{\scriptstyle2} \arrow[ur, mapsto, harpoon, "\textsf{head}"] \arrow[dr, mapsto, harpoon', "\textsf{tail\ }"'] & & {\scriptstyle2} \\[-2.5ex]
& {\scriptstyle5} \arrow[ur, mapsto] & 
\end{tikzcd}
\end{equation*}%

No: there exists a \texttt{Project} channel that undoes the action of \texttt{Product}.
Does it generate information?

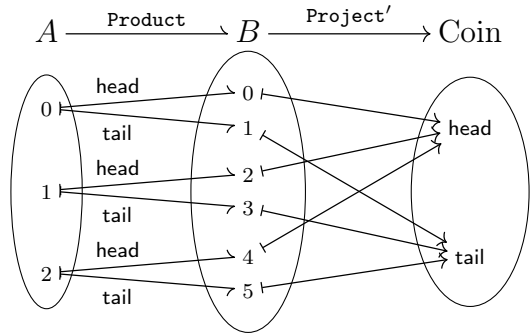
\begin{equation*}
\begin{tikzcd}[
    row sep=tiny, 
    column sep=huge,
    execute at end picture={
        \node[setellipse, fit={(\tikzcdmatrixname-3-1) (\tikzcdmatrixname-11-1)}, inner xsep=2pt, inner ysep=-7pt]{};
        \node[setellipse, fit={(\tikzcdmatrixname-2-2) (\tikzcdmatrixname-12-2)}, inner ysep=-7pt]{};
        \node[setellipse, fit={(\tikzcdmatrixname-4-3) (\tikzcdmatrixname-10-3)}, inner xsep=2pt]{};
    }
]
A \arrow[r, "\texttt{Product}", harpoon] & B \arrow[r, "\texttt{Project}'"] & \textup{Coin} \\[.5ex]
& {\scriptstyle0} \arrow[ddr, mapsto] & \\[-2.5ex]
{\scriptstyle0} \arrow[ur, mapsto, harpoon, "\textsf{head}"] \arrow[dr, mapsto, harpoon', "\textsf{tail\ }"'] & & \\[-2.5ex]
\phantom{\scriptstyle0} & {\scriptstyle1} \arrow[ddddddr, mapsto] & {\scriptstyle\textsf{head}} \\[-2.5ex]
\phantom{\scriptstyle0} & \\[-1.5ex]
\phantom{\scriptstyle0} & {\scriptstyle2} \arrow[uur, mapsto] \\[-2.5ex]
{\scriptstyle1} \arrow[ur, mapsto, harpoon, "\textsf{head}"] \arrow[dr, mapsto, harpoon', "\textsf{tail\ }"'] & &\\[-2.5ex]
\phantom{\scriptstyle0} & {\scriptstyle3} \arrow[ddr, mapsto] \\[-2.5ex]
\phantom{\scriptstyle0} & \\[-1.5ex]
\phantom{\scriptstyle0} & {\scriptstyle4} \arrow[uuuuuur, mapsto] & {\scriptstyle\textsf{tail}} \\[-2.5ex]
{\scriptstyle2} \arrow[ur, mapsto, harpoon, "\textsf{head}"] \arrow[dr, mapsto, harpoon', "\textsf{tail\ }"'] & & \\[-2.5ex]
& {\scriptstyle5} \arrow[uur, mapsto] & 
\end{tikzcd}
\end{equation*}%

Yes! Bob could use \texttt{Project'} to learn the outcome of the coin flip, an information that Alice is completely unaware of!

For our divide-and-conquer aims, we needed Bob to be easier to model than Alice, and we got the opposite: Bob is strictly harder to model than Alice because he has all of Alice's information, plus some irrelevant information about a coin.
Therefore, \texttt{Product} is an even worse channel than \texttt{Cypher} for this divide-and-conquer purpose.

If noise isn't what we need, what is it?

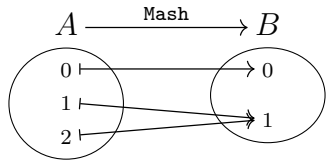
\begin{equation*}
\begin{tikzcd}[
    row sep=tiny, 
    column sep=huge,
    execute at end picture={
        \node[setellipse, fit={(\tikzcdmatrixname-2-1) (\tikzcdmatrixname-6-1)}, inner ysep=-5pt]{};
        \node[setellipse, fit={(\tikzcdmatrixname-2-2) (\tikzcdmatrixname-5-2)}, inner ysep=-4pt]{};
    }
]
A \arrow[r, "\texttt{Mash}"] & B \\[-.7ex]
{\scriptstyle0} \arrow[r, mapsto] & {\scriptstyle0} \\[-2.5ex]
\phantom{\scriptstyle0} & \\[-2.5ex]
{\scriptstyle1} \arrow[dr, mapsto] & \\[-2.5ex]
& {\scriptstyle1} \\[-2.5ex]
{\scriptstyle2} \arrow[ur, mapsto] &
\end{tikzcd}
\end{equation*}%

That's it!
When Bob gets the message $1$, he can't tell with certainty whether Alice said $1$ or $2$.
`Mash`ing different messages together is what destroys information, not noise.

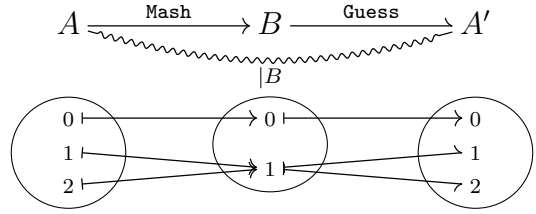
\begin{equation*}
\begin{tikzcd}[
    row sep=tiny, 
    column sep=huge,
    execute at end picture={
        \node[setellipse, fit={(\tikzcdmatrixname-2-1) (\tikzcdmatrixname-6-1)}, inner ysep=-5pt]{};
        \node[setellipse, fit={(\tikzcdmatrixname-2-2) (\tikzcdmatrixname-5-2)}, inner ysep=-4pt]{};
        \node[setellipse, fit={(\tikzcdmatrixname-2-3) (\tikzcdmatrixname-6-3)}, inner ysep=-5pt]{};
    }
]
A \arrow[r, "\texttt{Mash}"] \arrow[rr, no head, squigglish, bend right=15, "|B"'] & B \arrow[r, harpoon, "\texttt{Guess}"] & A' \\[3.0ex]
{\scriptstyle0} \arrow[r, mapsto] & {\scriptstyle0} \arrow[r, mapsto] & {\scriptstyle0} \\[-2.5ex]
\phantom{\scriptstyle0} & \\[-2.5ex]
{\scriptstyle1} \arrow[dr, mapsto] & & {\scriptstyle1}\\[-2.5ex]
& {\scriptstyle1} \arrow[ur, mapsto, harpoon] \arrow[dr, mapsto, harpoon'] \\[-2.5ex]
{\scriptstyle2} \arrow[ur, mapsto] & & {\scriptstyle2}
\end{tikzcd}
\end{equation*}%

When Bob gets the message $B=0$, he can simply assign $A'=0$.
But when $B=1$, he must learn to \texttt{Guess} the message sent by an Imaginary-Alice $A'$ so that $\textup{P}^\theta(A'|B)$ is as close as possible to $\textup{P}(A|B)$.

I indicate this last requirement using a ``squiggly'' line, with the conditional $|B$ annotating it.
However, in many applications, we don't have a particular desire for $\textup{P}^\theta(A'|B) = \textup{P}(A|B)$, and we could be perfectly content with any $\textup{P}^\theta(A'|B)$ such that $\textup{P}^\theta(A') = \textup{P}(A)$.
In those cases, I simply don't add any annotation to the squiggly line.

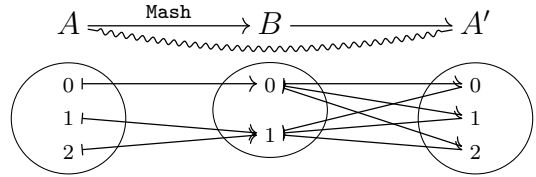
\begin{equation*}
\begin{tikzcd}[
    row sep=tiny, 
    column sep=huge,
    execute at end picture={
        \node[setellipse, fit={(\tikzcdmatrixname-2-1) (\tikzcdmatrixname-6-1)}, inner ysep=-5pt]{};
        \node[setellipse, fit={(\tikzcdmatrixname-2-2) (\tikzcdmatrixname-5-2)}, inner ysep=-4pt]{};
        \node[setellipse, fit={(\tikzcdmatrixname-2-3) (\tikzcdmatrixname-6-3)}, inner ysep=-5pt]{};
    }
]
A \arrow[r, "\texttt{Mash}"] \arrow[rr, no head, squigglish, bend right=10] & B \arrow[r, harpoon] & A' \\[.5ex]
{\scriptstyle0} \arrow[r, mapsto] & {\scriptstyle0} \arrow[r, mapsto, harpoon] \arrow[ddr, mapsto, harpoon] \arrow[ddddr, mapsto, harpoon] & {\scriptstyle0} \\[-2.5ex]
\phantom{\scriptstyle0} & \\[-2.5ex]
{\scriptstyle1} \arrow[dr, mapsto] & & {\scriptstyle1}\\[-2.5ex]
& {\scriptstyle1} \arrow[uuur, mapsto, harpoon] \arrow[ur, mapsto, harpoon] \arrow[dr, mapsto, harpoon] \\[-2.5ex]
{\scriptstyle2} \arrow[ur, mapsto] & & {\scriptstyle2}
\end{tikzcd}
\end{equation*}%

Note that I didn't name the arrow from Bob to the Imaginary-Alice, because I don't really care how Bob does it: as long as $\textup{P}^\theta(A') = \textup{P}(A)$, \emph{i.e.} the two people joined by an un-decorated squiggly line have the same marginal distribution, we're good.

And I may not even care about the details of the possible messages and how they relate, in which cases I simply ommit the ``set'' ellipses, the messages, and ``maps to'' arrows/harpoons between them.

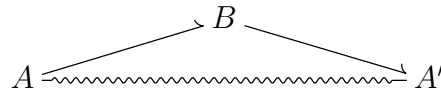
\begin{equation*}
\begin{tikzcd}[
    row sep=tiny, 
    column sep=huge
]
& B \arrow[dr, harpoon] & \\
A \arrow[ur, harpoon] \arrow[rr, no head, squigglish] & & A' 
\end{tikzcd}
\end{equation*}%

What does this picture tell us?
Well, for starters, the squiggly line tells us that $A$ and $A'$ have the same marginal probability distribution, so they must have the same information content (entropy)

$$
\textup{H}(A) = \textup{H}(A') .
$$

How does $\textup{H}(B)$ relate to those two?
We don't know: it could be higher, lower, or equal, depending on the details of those communication channels.
One thing is clear though: as much information must be created/destroyed from $A$ to $B$ as is destroyed/created from $B$ to $A'$.

Ok, so let's drop the squiggly line and see what else the diagram says.

\begin{equation*}
\begin{tikzcd}[
    row sep=tiny, 
    column sep=huge
]
A \arrow[r, harpoon] & B \arrow[r, harpoon] & A' 
\end{tikzcd}
\end{equation*}%

This picture can be understood in the context of the data processing inequality.
From~\textcite{Latham2009}:

\begin{blockquote}
The Data Processing Inequality (DPI) states, loosely, that post-processing cannot increase information.
\end{blockquote}

If you think ``Haven't we just shown that communication channels may create information?!'', then yes, you are correct.
But the information created by communication channels \emph{is completely irrelevant to anything that came before in the communication chain}.
Latham and Roudi are saying that information about an early message cannot appear in later messages by further processing.
This ``information about'' can be measured with the mutual information; we write $\textup{I}(Y; X)$ the mutual information between $X$ and $Y$, and it satisfies

$$
\textup{I}(Y; X) = \textup{H}(Y) - \textup{H}(X | Y) = \textup{H}(X) - \textup{H}(Y | X)  , \qquad \textup{I}(Y; Y) = \textup{H}(Y) .
$$

Therefore, what the data processing inequality tells us is that

\begin{align*}
\textup{I}(A; A) \ge \textup{I}(A; B) & \ge \textup{I}(A; A') \cr
\textup{I}(B; B) & \ge \textup{I}(B; A') .
\end{align*}

Whatever there is to know about $A$, $B$ cannot know more than that, and $A'$ cannot know more than $B$ did.
And $A'$ cannot know more about $B$ than $B$ knew about himself.

This is what I call ``the destruction story'': \textbf{you cannot know what has been forgotten before you heard about it.}
However, the data processing inequality has a lesser known dual, which I call ``the generation story''

\begin{align*}
\textup{I}(A'; A) & \le \textup{I}(A'; B) \le \textup{I}(A'; A') \cr
\textup{I}(B; A) & \le \textup{I}(B; B) .
\end{align*}

$A$ cannot know more than $B$ about $A'$, and $B$ cannot know more about $A'$ than there is to be known about $A'$.
And $A$ cannot know more than $B$ about $B$.
In the generation story, \textbf{you cannot know about what has not been decided yet.}

These zoomed-out diagrams are a great way to consider general classes of problems.
But ultimately, the generation/destruction of information happens at the zoomed-in, message level.

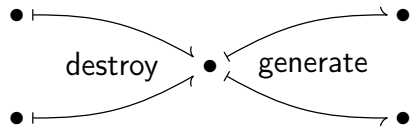
\begin{equation*}
\begin{tikzcd}[
    row sep=tiny, 
    column sep=huge
]
\bullet \arrow[dr, mapsto, harpoon, bend left=15] && \bullet \\
\ \arrow[r, phantom, "\textsf{destroy}"] & \bullet \arrow[ur, mapsto, harpoon, bend left=15] \arrow[dr, mapsto, harpoon', bend right=15]  \arrow[r, phantom, "\textsf{  generate}"] & \ \\
\bullet \arrow[ur, mapsto, harpoon', bend right=15] && \bullet
\end{tikzcd}
\end{equation*}%

\begin{itemize}
\item When different messages converge to the same message, information is \textbf{destroyed}.
\item When a given message can diverge into different messages, information is \textbf{generated}.
\end{itemize}

\subsection{Mashing Everything}

Let's continue from the \texttt{Mash} example in the previous section.
Now suppose that Bob passes the message to Carol through the \texttt{Mash'} channel.

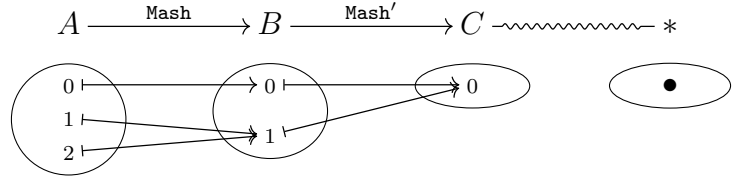
\begin{equation*}
\begin{tikzcd}[
    row sep=tiny, 
    column sep=huge,
    execute at end picture={
        \node[setellipse, fit={(\tikzcdmatrixname-2-1) (\tikzcdmatrixname-6-1)}, inner ysep=-5pt]{};
        \node[setellipse, fit={(\tikzcdmatrixname-2-2) (\tikzcdmatrixname-5-2)}, inner ysep=-4pt]{};
        \node[setellipse, fit={(\tikzcdmatrixname-2-3) (\tikzcdmatrixname-2-3)}]{};
        \node[setellipse, fit={(\tikzcdmatrixname-2-4) (\tikzcdmatrixname-2-4)}]{};
    }
]
A \arrow[r, "\texttt{Mash}"] & B \arrow[r, "\texttt{Mash}'"] & C \arrow[r, no head, squigglish] & * \\[.5ex]
{\scriptstyle0} \arrow[r, mapsto] & {\scriptstyle0} \arrow[r, mapsto] & {\scriptstyle0} & \bullet \\[-2.5ex]
\phantom{\scriptstyle0} & \\[-2.5ex]
{\scriptstyle1} \arrow[dr, mapsto]\\[-2.5ex]
& {\scriptstyle1} \arrow[uuur, mapsto] \\[-2.5ex]
{\scriptstyle2} \arrow[ur, mapsto] &
\end{tikzcd}
\end{equation*}%

Notice that Carol always gets the same message, which is called a singleton: I highlight this by linking her to a special ``*'' symbol using a squiggly line.
All singleton have no information: $\textup{H}(C) = 0$.

But by the definition of mutual information, we must have $\textup{I}(A; C) \le \textup{H}(C) = 0$: Carol cannot know more about Alice than Carol knows at all, but Carol knows nothing, therefore Carol knows nothing about Alice.
Stated otherwise, all information about Alice has been destroyed by the time the messages get to Carol.
Success!
We have divided a large task into smaller, simpler ones, until all that was left was trivial!

Now unto the ``conquer'' part of divide-and-conquer.

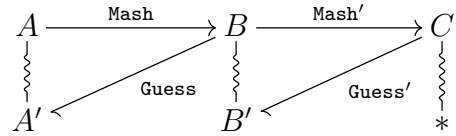
\begin{equation*}
\begin{tikzcd}[
    row sep=tiny, 
    column sep=huge,
]
A \arrow[r, "\texttt{Mash}"] & B \arrow[r, "\texttt{Mash}'"] \arrow[dl, harpoon, "\texttt{Guess}"] & C \arrow[dl, harpoon, "\texttt{Guess}'"] \\[2.5ex]
A' \arrow[u, no head, squigglish] & B' \arrow[u, no head, squigglish] & * \arrow[u, no head, squigglish]
\end{tikzcd}
\end{equation*}%

Bob can learn \texttt{Guess} using \texttt{Mash} on Alice's message, and Carol can learn \texttt{Guess'} using \texttt{Mash'} on Bob's.
And because Carol knows nothing, we're ready to generate completely fake $A''$ behavior from nothing! 

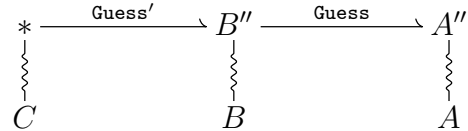
\begin{equation*}
\begin{tikzcd}[
    row sep=tiny, 
    column sep=huge,
]
* \arrow[r, harpoon, "\texttt{Guess}'"] & B'' \arrow[r, harpoon, "\texttt{Guess}"] & A'' \\[2.5ex]
C \arrow[u, no head, squigglish] & B \arrow[u, no head, squigglish] & A \arrow[u, no head, squigglish]
\end{tikzcd}
\end{equation*}%

This is the strategy used in autoregressive language models.

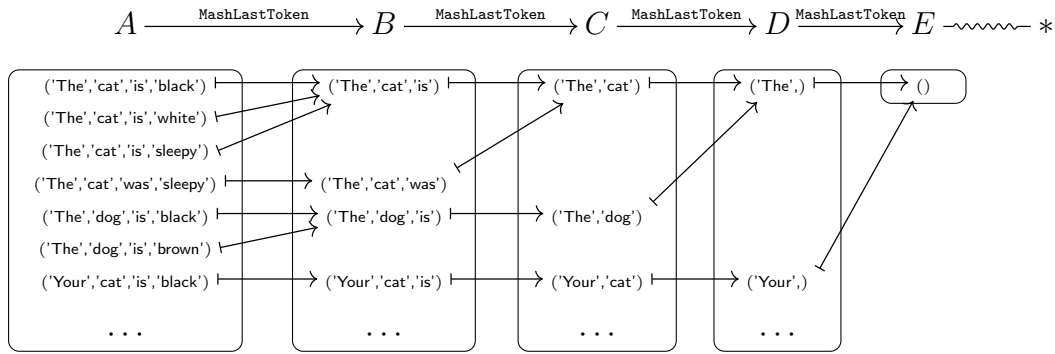
\begin{equation*}
\begin{tikzcd}[
    row sep=tiny, 
    execute at end picture={
        \node[setbox, fit={(\tikzcdmatrixname-2-1) (\tikzcdmatrixname-9-1)}]{};
        \node[setbox, fit={(\tikzcdmatrixname-2-2) (\tikzcdmatrixname-9-2)}]{};
        \node[setbox, fit={(\tikzcdmatrixname-2-3) (\tikzcdmatrixname-9-3)}]{};
        \node[setbox, fit={(\tikzcdmatrixname-2-4) (\tikzcdmatrixname-9-4)}]{};
        \node[setbox, fit={(\tikzcdmatrixname-2-5) (\tikzcdmatrixname-2-5)}]{};
    }
]
A \arrow[r, "{\scriptscriptstyle\texttt{MashLastToken}}"] & B \arrow[r, "{\scriptscriptstyle\texttt{MashLastToken}}"] & C\arrow[r, "{\scriptscriptstyle\texttt{MashLastToken}}"] & D\arrow[r, "{\scriptscriptstyle\texttt{MashLastToken}}"] & E \arrow[r, no head, squigglish] & * \\[.5ex]
{\scriptscriptstyle(\textsf{'The','cat','is','black'})} \arrow[r, mapsto] & 
{\scriptscriptstyle(\textsf{'The','cat','is'})} \arrow[r, mapsto] & 
{\scriptscriptstyle(\textsf{'The','cat'})} \arrow[r, mapsto] & 
{\scriptscriptstyle(\textsf{'The',})} \arrow[r, mapsto] &
{\scriptscriptstyle()} \\[-1.5ex]
{\scriptscriptstyle(\textsf{'The','cat','is','white'})} \arrow[ur, mapsto, start anchor=east] \\[-1.5ex]
{\scriptscriptstyle(\textsf{'The','cat','is','sleepy'})} \arrow[uur, mapsto, start anchor=east] \\[-1.5ex]
{\scriptscriptstyle(\textsf{'The','cat','was','sleepy'})} \arrow[r, mapsto, start anchor=east] &
{\scriptscriptstyle(\textsf{'The','cat','was'})} \arrow[uuur, mapsto, start anchor=north east] \\[-1.5ex]
{\scriptscriptstyle(\textsf{'The','dog','is','black'})} \arrow[r, mapsto, start anchor=east] &
{\scriptscriptstyle(\textsf{'The','dog','is'})} \arrow[r, mapsto, start anchor=east] &
{\scriptscriptstyle(\textsf{'The','dog'})} \arrow[uuuur, mapsto, start anchor=north east] \\[-1.5ex]
{\scriptscriptstyle(\textsf{'The','dog','is','brown'})} \arrow[ur, mapsto, start anchor=east] \\[-1.5ex]
{\scriptscriptstyle(\textsf{'Your','cat','is','black'})} \arrow[r, mapsto, start anchor=east] &
{\scriptscriptstyle(\textsf{'Your','cat','is'})} \arrow[r, mapsto, start anchor=east] &
{\scriptscriptstyle(\textsf{'Your','cat'})} \arrow[r, mapsto, start anchor=east] &
{\scriptscriptstyle(\textsf{'Your',})} \arrow[uuuuuur, mapsto, start anchor=north east] \\
\cdots & \cdots & \cdots & \cdots
\end{tikzcd}
\end{equation*}%

Here the ``$\cdots$'' represent additional messages whose arrows are not explicitly shown.
The message is iteratively fed through the \texttt{MashLastToken} communication channel, which mashes together all the messages that have the same prefix up to the last token.
Because of the setting's symmetric structure, we can use the same neural network at each position to predict the probability distribution for what the next should be: all positions can contribute to train the same \texttt{GuessNextToken}.
At inference, we start from the singleton (the empty sequence), then iteratively apply \texttt{GuessNextToken}.
Et voilà!
An autoregressive language model!

Notice how each time we destroy some information, we're carving out a chunk of the overall task to be learned.
Here divide-and-conquer amounted to ``learn to generate one token at a time''.
Such a very structured, regular way to destroy the message, is typical of many non-diffusion machine learning techniques.
One of my claims in this blogpost is that diffusion models can destroy information more ``organically'', possibly reducing the human-designer bias hand-waved at by~\textcite{sutton2019bitter}
As an example of what I mean, here's what the last diagram would look like for a mask diffusion language model.

{\tiny
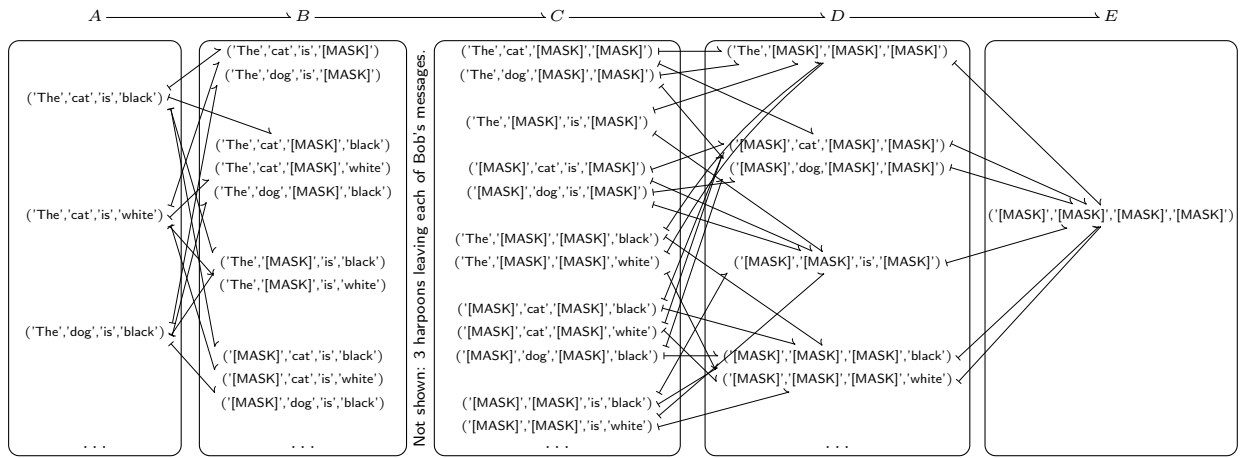
\begin{equation*}
\begin{tikzcd}[
    row sep=tiny,
    execute at end picture={
        \node[setbox, fit={(\tikzcdmatrixname-2-1) (\tikzcdmatrixname-19-1)}, inner xsep=25pt]{};
        \node[setbox, fit={(\tikzcdmatrixname-2-2) (\tikzcdmatrixname-19-2)}, inner xsep=7pt] (boxB) {};
        \node[setbox, fit={(\tikzcdmatrixname-2-3) (\tikzcdmatrixname-19-3)}, inner xsep=7pt] (boxC) {};
        \node[setbox, fit={(\tikzcdmatrixname-2-4) (\tikzcdmatrixname-19-4)}, inner xsep=5pt]{};
        \node[setbox, fit={(\tikzcdmatrixname-2-5) (\tikzcdmatrixname-19-5)}, inner xsep=40pt]{};
        \node[rotate=90, font=\tiny\sffamily, anchor=center]
        at ($(boxB.east)!0.51!(boxC.west)$)
        {Not shown: 3 harpoons leaving each of Bob's messages.};
    }
]
A \arrow[r, harpoon] &[-5pt] B \arrow[r, harpoon] & C\arrow[r, harpoon] & D\arrow[r, harpoon] &[-10pt] E \\[.5ex]
\phantom{\scriptscriptstyle()} & {\scriptscriptstyle(\textsf{'The','cat','is','[MASK]'})} & {\scriptscriptstyle(\textsf{'The','cat','[MASK]','[MASK]'})} \arrow[r, mapsto, harpoon, start anchor=east] \arrow[ddddr, mapsto, harpoon, start anchor=south east] & {\scriptscriptstyle(\textsf{'The','[MASK]','[MASK]','[MASK]'})} \arrow[dddddddr, mapsto, harpoon, start anchor=south east] & \phantom{\scriptscriptstyle()} \\[-1.5ex]
& {\scriptscriptstyle(\textsf{'The','dog','is','[MASK]'})} & {\scriptscriptstyle(\textsf{'The','dog','[MASK]','[MASK]'})} \arrow[ur, mapsto, harpoon, start anchor=east] \arrow[ddddr, mapsto, harpoon, start anchor=south east, end anchor=north west] \\[-1.5ex]
{\scriptscriptstyle(\textsf{'The','cat','is','black'})} \arrow[uur, mapsto, harpoon, start anchor=north east, end anchor=west] \arrow[ddr, mapsto, harpoon, start anchor=east] \arrow[dddddddr, mapsto, harpoon, start anchor=south east, end anchor=north west] \arrow[dddddddddddr, mapsto, harpoon, start anchor=south east, end anchor=north west] & \\[-1.5ex]
\ & & {\scriptscriptstyle(\textsf{'The','[MASK]','is','[MASK]'})} \arrow[uuur, mapsto, harpoon, start anchor=north east] \arrow[ddddddr, mapsto, harpoon, start anchor=south east] \\[-1.5ex]
& {\scriptscriptstyle(\textsf{'The','cat','[MASK]','black'})} & & {\scriptscriptstyle(\textsf{'[MASK]','cat','[MASK]','[MASK]'})} \arrow[dddr, mapsto, harpoon, start anchor=east] \\[-1.5ex]
& {\scriptscriptstyle(\textsf{'The','cat','[MASK]','white'})} & {\scriptscriptstyle(\textsf{'[MASK]','cat','is','[MASK]'})} \arrow[ur, mapsto, harpoon, start anchor=east, end anchor=west] \arrow[ddddr, mapsto, harpoon, start anchor=south east] & {\scriptscriptstyle(\textsf{'[MASK]','dog,'[MASK]','[MASK]'})} \arrow[ddr, mapsto, harpoon, start anchor=east] \\[-1.5ex]
& {\scriptscriptstyle(\textsf{'The','dog','[MASK]','black'})} & {\scriptscriptstyle(\textsf{'[MASK]','dog','is','[MASK]'})} \arrow[ur, mapsto, harpoon, start anchor=east] \arrow[dddr, mapsto, harpoon, start anchor=south east] \\[-1.5ex]
{\scriptscriptstyle(\textsf{'The','cat','is','white'})} \arrow[uuuuuuur, mapsto, harpoon, start anchor=north east, end anchor=south west] \arrow[uur, mapsto, harpoon, start anchor=east, end anchor=south west] \arrow[dddr, mapsto, harpoon, start anchor=south east, end anchor=north west] \arrow[dddddddr, mapsto, harpoon, start anchor=south east, end anchor=north west] & & & & {\scriptscriptstyle(\textsf{'[MASK]','[MASK]','[MASK]','[MASK]'})} \\[-1.5ex]
\ & & {\scriptscriptstyle(\textsf{'The','[MASK]','[MASK]','black'})} \arrow[uuuuuuuur, mapsto, harpoon, start anchor=north east, bend left=10] \arrow[dddddr, mapsto, harpoon, start anchor=east] \\[-1.5ex]
& {\scriptscriptstyle(\textsf{'The','[MASK]','is','black'})} & {\scriptscriptstyle(\textsf{'The','[MASK]','[MASK]','white'})} \arrow[uuuuuuuuur, mapsto, harpoon, start anchor=north east, bend left=10] \arrow[dddddr, mapsto, harpoon, start anchor=south east, end anchor=north west] & {\scriptscriptstyle(\textsf{'[MASK]','[MASK]','is','[MASK]'})} \arrow[uur, mapsto, harpoon, start anchor=east] \\[-1.5ex]
& {\scriptscriptstyle(\textsf{'The','[MASK]','is','white'})} \\[-1.5ex]
\ &  & {\scriptscriptstyle(\textsf{'[MASK]','cat','[MASK]','black'})} \arrow[uuuuuuur, mapsto, harpoon, start anchor=north east, end anchor=south west] \arrow[ddr, mapsto, harpoon, start anchor=east] \\[-1.5ex]
{\scriptscriptstyle(\textsf{'The','dog','is','black'})} \arrow[uuuuuuuuuuur, mapsto, harpoon, start anchor=north east, end anchor=south west] \arrow[uuuuuur, mapsto, harpoon, start anchor=east, end anchor=south west] \arrow[uuur, mapsto, harpoon, start anchor=east, end anchor=south west] \arrow[dddr, mapsto, harpoon, start anchor=south east, end anchor=north west] & & {\scriptscriptstyle(\textsf{'[MASK]','cat','[MASK]','white'})} \arrow[uuuuuuuur, mapsto, harpoon, start anchor=north east, end anchor=south west] \arrow[ddr, mapsto, harpoon, start anchor=east, end anchor=west] \\[-1.5ex]
 & {\scriptscriptstyle(\textsf{'[MASK]','cat','is','black'})} & {\scriptscriptstyle(\textsf{'[MASK]','dog','[MASK]','black'})} \arrow[uuuuuuuur, mapsto, harpoon, start anchor=north east, end anchor=south west] \arrow[r, mapsto, harpoon]  & {\scriptscriptstyle(\textsf{'[MASK]','[MASK]','[MASK]','black'})} \arrow[uuuuuur, mapsto, harpoon, start anchor=east] \\[-1.5ex]
& {\scriptscriptstyle(\textsf{'[MASK]','cat','is','white'})} & & {\scriptscriptstyle(\textsf{'[MASK]','[MASK]','[MASK]','white'})} \arrow[uuuuuuur, mapsto, harpoon, start anchor=east] \\[-1.5ex]
& {\scriptscriptstyle(\textsf{'[MASK]','dog','is','black'})} & {\scriptscriptstyle(\textsf{'[MASK]','[MASK]','is','black'})} \arrow[uuuuuur, mapsto, harpoon, start anchor=north east, end anchor=south west] \arrow[uur, mapsto, harpoon, start anchor=east, end anchor=south west]  \\[-1.5ex]
\ & & {\scriptscriptstyle(\textsf{'[MASK]','[MASK]','is','white'})} \arrow[uuuuuuur, mapsto, harpoon, start anchor=north east] \arrow[uur, mapsto, harpoon, start anchor=east]  \\[-1.5ex]
\cdots & \cdots & \cdots & \cdots & \phantom{\cdots}
\end{tikzcd}
\end{equation*}}%

(There should be 3 arrows leaving each of Bob's messages toward Carol, please pardon me for eschewing them.)

While \texttt{MashLastToken} was strictly destroying information, \texttt{MashOneRandomUnmaskedToken} both generates information (which token is to be masked) and destroys information (mash together all messages that would be the same if it were not for the randomly-selected unmasked token).
However, the newly generated information (masking order) is also destroyed by the process.
In the end, we still get a singleton: all information is eventually destroyed.

The point I'm trying to make here is that \texttt{MashOneRandomUnmaskedToken} is a richer approach to ``divide and conquer'' than \texttt{MashLastToken} was.
It is more ``organic'' in a similar sense to how mixing milk in coffee is ``organic'': it's a mess, a \emph{rich} mess.\footnote{Further notice that this diagram used less Alice messages than the one I showed in the autoregressive case: the picture is a bigger mess despite me showing less details.}
We can train a model to \texttt{GuessOneRandomMaskToken} on that mess, and yes, I could excuse it to train slower than \texttt{GuessNextToken}, because \texttt{GuessNextToken} is a subset of what \texttt{GuessOneRandomMaskToken} is learning.
Let me make it explicit.

{\tiny
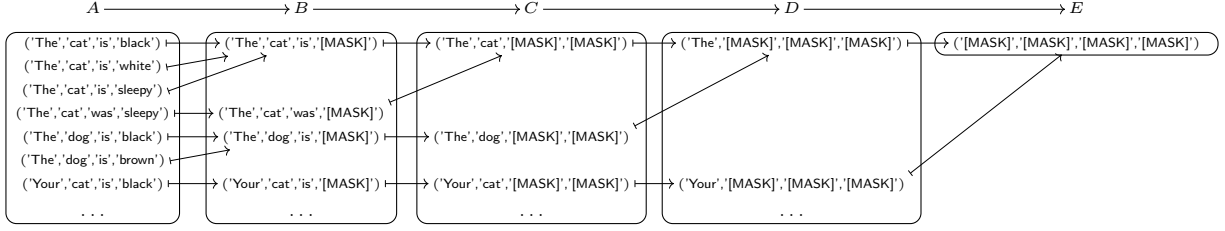
\begin{equation*}
\begin{tikzcd}[
    row sep=tiny, 
    execute at end picture={
        \node[setbox, fit={(\tikzcdmatrixname-2-1) (\tikzcdmatrixname-9-1)}]{};
        \node[setbox, fit={(\tikzcdmatrixname-2-2) (\tikzcdmatrixname-9-2)}]{};
        \node[setbox, fit={(\tikzcdmatrixname-2-3) (\tikzcdmatrixname-9-3)}]{};
        \node[setbox, fit={(\tikzcdmatrixname-2-4) (\tikzcdmatrixname-9-4)}]{};
        \node[setbox, fit={(\tikzcdmatrixname-2-5) (\tikzcdmatrixname-2-5)}]{};
    }
]
A \arrow[r] &[-5pt] B \arrow[r] &[-5pt] C\arrow[r] &[-5pt] D\arrow[r] &[-5pt] E \\[.5ex]
{\scriptscriptstyle(\textsf{'The','cat','is','black'})} \arrow[r, mapsto] & 
{\scriptscriptstyle(\textsf{'The','cat','is','[MASK]'})} \arrow[r, mapsto] & 
{\scriptscriptstyle(\textsf{'The','cat','[MASK]','[MASK]'})} \arrow[r, mapsto] & 
{\scriptscriptstyle(\textsf{'The','[MASK]','[MASK]','[MASK]'})} \arrow[r, mapsto] &
{\scriptscriptstyle(\textsf{'[MASK]','[MASK]','[MASK]','[MASK]'})} \\[-1.5ex]
{\scriptscriptstyle(\textsf{'The','cat','is','white'})} \arrow[ur, mapsto, start anchor=east] \\[-1.5ex]
{\scriptscriptstyle(\textsf{'The','cat','is','sleepy'})} \arrow[uur, mapsto, start anchor=east] \\[-1.5ex]
{\scriptscriptstyle(\textsf{'The','cat','was','sleepy'})} \arrow[r, mapsto, start anchor=east] &
{\scriptscriptstyle(\textsf{'The','cat','was','[MASK]'})} \arrow[uuur, mapsto, start anchor=north east] \\[-1.5ex]
{\scriptscriptstyle(\textsf{'The','dog','is','black'})} \arrow[r, mapsto, start anchor=east] &
{\scriptscriptstyle(\textsf{'The','dog','is','[MASK]'})} \arrow[r, mapsto, start anchor=east] &
{\scriptscriptstyle(\textsf{'The','dog','[MASK]','[MASK]'})} \arrow[uuuur, mapsto, start anchor=north east] \\[-1.5ex]
{\scriptscriptstyle(\textsf{'The','dog','is','brown'})} \arrow[ur, mapsto, start anchor=east] \\[-1.5ex]
{\scriptscriptstyle(\textsf{'Your','cat','is','black'})} \arrow[r, mapsto, start anchor=east] &
{\scriptscriptstyle(\textsf{'Your','cat','is','[MASK]'})} \arrow[r, mapsto, start anchor=east] &
{\scriptscriptstyle(\textsf{'Your','cat','[MASK]','[MASK]'})} \arrow[r, mapsto, start anchor=east] &
{\scriptscriptstyle(\textsf{'Your','[MASK]','[MASK]','[MASK]'})} \arrow[uuuuuur, mapsto, start anchor=north east] \\
\cdots & \cdots & \cdots & \cdots
\end{tikzcd}
\end{equation*}}%

We could train a model to \texttt{GuessOneRandomMaskToken} then, at inference, use the learned weight to perform the \texttt{GuessNextToken} task instead.
In the scenario where we're data-starved, but we can train for as long as we wish to, with a model as big as it needs to~\autocite{ni2025diffusion}, and inference is performed using \texttt{GuessNextToken}, which model would you put your money on: the one trained solely on this specific \texttt{GuessNextToken} task, or the one that saw the worst~\autocite{kim2025train} that \texttt{GuessOneRandomMaskToken} had to offer?

\subsection{Shuffling Everything}

In the previous section, we saw that both autoregressive language models and mask diffusion language models learn to mimic Alice's message distribution by iteratively \texttt{Mash}ing them toward a singleton, reaching a state with no information about anything, thus no information about Alice.
Here we consider the other leading approach to destroying Alice's message: drowning it into irrelevant information.

This mechanism, which I call \texttt{Shuffle}, may align better with the historic/physical meaning of the word ``diffusion''.\footnote{In other words, while I expect that some readers could say that \texttt{Mash} isn't ``real'' diffusion, I don't have the same worry for \texttt{Shuffle}.}
Whereas \texttt{Mash} \emph{may} generate information (\emph{e.g}., masking order) but will eventually destroy everything to a singleton, \texttt{Shuffle} \emph{must} generate new information to be folded into Alice's message, drowning it in the noise.

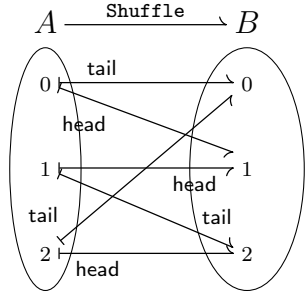
\begin{equation*}
\begin{tikzcd}[
    row sep=tiny, 
    column sep=huge,
    execute at end picture={
        \node[setellipse, fit={(\tikzcdmatrixname-2-1) (\tikzcdmatrixname-4-1)}, inner xsep=2pt, inner ysep=-7pt]{};
        \node[setellipse, fit={(\tikzcdmatrixname-2-2) (\tikzcdmatrixname-4-2)}, inner ysep=-7pt]{};
    }
]
A \arrow[r, "\texttt{Shuffle}", harpoon] & B \\[.5ex]
{\scriptstyle0} \arrow[r, mapsto, harpoon, "\textsf{tail\ }", near start] \arrow[dr, mapsto, harpoon', "\textsf{head}"', pos=0.3, end anchor=north west] & {\scriptstyle0} \\[2.5ex]
{\scriptstyle1} \arrow[r, mapsto, harpoon', "\textsf{head}"', pos=0.8] \arrow[dr, mapsto, harpoon, "\textsf{tail\ }", pos=0.8] & {\scriptstyle1} \\[2.5ex]
{\scriptstyle2} \arrow[r, mapsto, harpoon, "\textsf{head}"', pos=0.2] \arrow[uur, mapsto, harpoon', "\textsf{tail\ }", pos=0.05] & {\scriptstyle2}
\end{tikzcd}
\end{equation*}%

If Bob has a good understanding of the \texttt{Shuffle} channel, he can model the situation as

$$
\begin{bmatrix}
    \textup{P}(B=0) \cr
    \textup{P}(B=1) \cr
    \textup{P}(B=2)
\end{bmatrix} =
\begin{bmatrix}
    1/2 & 0   & 1/2 \cr
    1/2 & 1/2 & 0   \cr
    0   & 1/2 & 1/2
\end{bmatrix}
\begin{bmatrix}
    \textup{P}(A=0) \cr
    \textup{P}(A=1) \cr
    \textup{P}(A=2)
\end{bmatrix} .
$$

When he observes $B=0$, he knows that Alice said either $0$ or $2$, but he's not sure which: information was destroyed, and he may learn to to generate $A'$, seeking to capture Alice's marginal probability distribution $\textup{P}(A)$.

But notice that (irrelevant) information was also created: when $B=0$, a Bob that understands the channel knows with certainty that the coin flip was ``tail'' (whereas this coin information was completely unknown to Alice).
In my view, \texttt{Shuffle} can destroy information about the original message \emph{because} it generates irrelevant information and has to fold the outcome in as many possible messages than there were before.
The message $A$ can take 3 values, the coin can take two values, the product of the coin and the message can take 6 values, and somehow this has to be crammed into the 3 possible values for $B$.
Something has to give; information about Alice is destroyed.

Next Bob can message Carol through the same \texttt{Shuffle} channel, and the same procedure can continue up to Zalgo.

$$
\begin{bmatrix}
    \textup{P}(Z=0) \cr
    \textup{P}(Z=1) \cr
    \textup{P}(Z=2)
\end{bmatrix} =
\begin{bmatrix}
    1/2 & 0   & 1/2 \cr
    1/2 & 1/2 & 0   \cr
    0   & 1/2 & 1/2
\end{bmatrix}^{25}
\begin{bmatrix}
    \textup{P}(A=0) \cr
    \textup{P}(A=1) \cr
    \textup{P}(A=2)
\end{bmatrix}
$$

Now, technically, the probability distribution $\textup{P}(Z)$ still depends on $\textup{P}(A)$, but what does that mean in practice?
Well, it can be shown that

$$
0.33333331 < P(Z=z) < 0.33333335 \quad \forall z \in \{0,1,2\} ,
$$

irrespective of Alice's probability distribution.
The crux of the \texttt{Shuffle} learning strategy here amounts to approximating Zalgo as a uniform distribution: 

$$
P(Z'=z') = \frac{1}{3} \quad \forall z' \in \{ 0,1,2 \}.
$$

So, whereas \texttt{Mash} eventually takes us to a singleton, \texttt{Shuffle} takes us to a known distribution, simple to sample.
Except for this detail, inference for \texttt{Shuffle} proceeds the same way as \texttt{Mash}: we sample $Z'$ from the known distribution, then apply the respective learned \texttt{Guess} down the alphabet until we get $A'$.

Like the previous section adapted the spirit of \texttt{Mash} to \texttt{MashOneRandomUnmaskedToken}, we can adapt \texttt{Shuffle} to obtain \texttt{ShuffleOneRandomToken}: a random token position gets substituted by a token selected uniformly at random from the tokenizer's vocabulary.
This is basically the ``Uniform'' noising process used in SEDD~\autocite{lou2024discrete}.

For the rest of this section, let's quickly consider what ``shuffling'' means in continuous space.
Suppose Alice's message is a $w \times h \times 3$ tensor of real numbers in the $[-1,1]$ interval representing an RGB picture of $w$ pixels wide by $h$ pixels high.
We may define the \texttt{GaussianShuffle($\delta$)} communication channel such that it adds independent Gaussian noise (with mean $0$ and variance $0 < \delta \ll 1$) to each entry of this tensor.

This channel generates (irrelevant) information: for a given message from Alice, there are many possible options for what Bob may receive; arrows branching out mean that information is generated.
But the channel also destroys information: there are many messages that Alice could have said that may explain a given message received by Bob; arrows converging in mean that information is destroyed.
As for \texttt{Shuffle}, \texttt{GaussianShuffle($\delta$)} destroys information about Alice's message by generating irrelevant information and folding it into limited space.\footnote{The story is a bit more complex here because, in a strict mathematical sense, specifying a single real number requires an infinite amount of information. One can work around this by considering differential entropy instead of entropy. However, for practical computer science applications, there is a much simpler resolution: these ``real'' numbers are represented as discrete data types. Concretely, a \texttt{float32} is really a discrete variable that may take one of $2^{32}$ values, thus capturing at most 32 bits of information.}

The impact of $n$ repeated independent applications of \texttt{GaussianShuffle($\delta$)} is a single ``bigger'' \texttt{GaussianShuffle(n*$\delta$)}.\footnote{Recall that the convolution of two Gaussians is a Gaussian whose mean and variance is the sum of their respective means and variances.}
Similarly to how we approximated Zalgo as a uniform distribution earlier, we can choose an $n$ that is high enough so that the outcome of applying \texttt{GaussianShuffle(n*$\delta$)} on any image is basically an $w \times h \times 3$ Gaussian with mean zero and variance $n\delta$, \emph{i.e.} no dependency worth mentioning on the actual image.
In essence, this amounts to the ``Variance Exploding'' process from~\textcite{song2021scorebased}; the ``Variance Preserving'' version could be similarly obtained by appropriately scaling down the outcome after each addition of Gaussian noise.

\subsection{Beyond Markov}

In our quest to divide-and-conquer the learning of Alice's distribution, we have seen two different strategies to destroy all the information in her message: \texttt{Mash} everything into a singleton, and \texttt{Shuffle} long enough to get a maximum-entropy distribution (\emph{e.g.} uniform or Gaussian).

Both these strategies are Markovian: each person's message has all the information required to get the probability distribution for the next person's message.
Stated differently, the ``spine'' of all my diagrams looked like this.

\begin{equation*}
\begin{tikzcd}[
    row sep=tiny, 
    column sep=large,
]
A \arrow[r, harpoon] & B \arrow[r, harpoon] & C \arrow[r, harpoon] & \cdots \arrow[r, harpoon] & X \arrow[r, harpoon] & Y \arrow[r, harpoon] & Z
\end{tikzcd}
\end{equation*}%

But this is not the only possibility, and many diffusion models have a non-Markovian approach to destruction and/or generation.
For example, although DDPM's~\autocite{ho2020denoising} destroying process may be framed as a Markovian chain, we may also frame it as interpolating between a ``clean'' sample $A'$ and a ``noise distribution'' $Z$

$$
B' = \sqrt{\alpha_B} A'  + \sqrt{1 - \alpha_B} Z', \quad C' = \sqrt{\alpha_C} A'  + \sqrt{1 - \alpha_C} Z', \quad \cdots \ ,
$$

given appropriate $1 = \alpha_A \ge \alpha_B \ge \alpha_C \ge \cdots \ge \alpha_Z = 0$.
Below is a diagram showing the Markovian approach on the top and the non-Markovian one on the bottom.

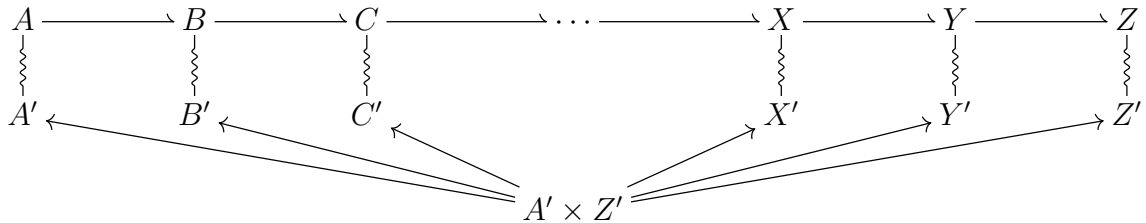
\begin{equation*}
\begin{tikzcd}[
    row sep=tiny, 
    column sep=large,
]
A \arrow[r, harpoon] & B \arrow[r, harpoon] & C \arrow[r, harpoon] & \cdots \arrow[r, harpoon] & X \arrow[r, harpoon] & Y \arrow[r, harpoon] & Z \\[2.5ex]
A' \arrow[u, no head, squigglish] & B' \arrow[u, no head, squigglish] & C' \arrow[u, no head, squigglish] & & X'  \arrow[u, no head, squigglish] & Y'\arrow[u, no head, squigglish] & Z' \arrow[u, no head, squigglish] \\[2.5ex]
&&& A' \times Z' \arrow[ulll] \arrow[ull] \arrow[ul] \arrow[ur] \arrow[urr] \arrow[urrr]
\end{tikzcd}
\end{equation*}%

Here I used $A' \times Z'$ to denote someone that has simultaneous access to both $A'$ and $Z'$ messages, and is thus able to use the previous equation to obtain $B'$, $C'$, \emph{etc.}

But once you've used $A' \times Z'$ to calculate, say, $B'$, you could just \emph{not discard} $Z'$, and obtain $B' \times Z'$.
The same holds up to $Y' \times Z'$ and, with a little more algebra, we can actually go back to $A' \times Z'$ (ignoring finite-precision errors).
I represent such information-preserving back-and-forth conversions using bidirectional arrows.

{\small
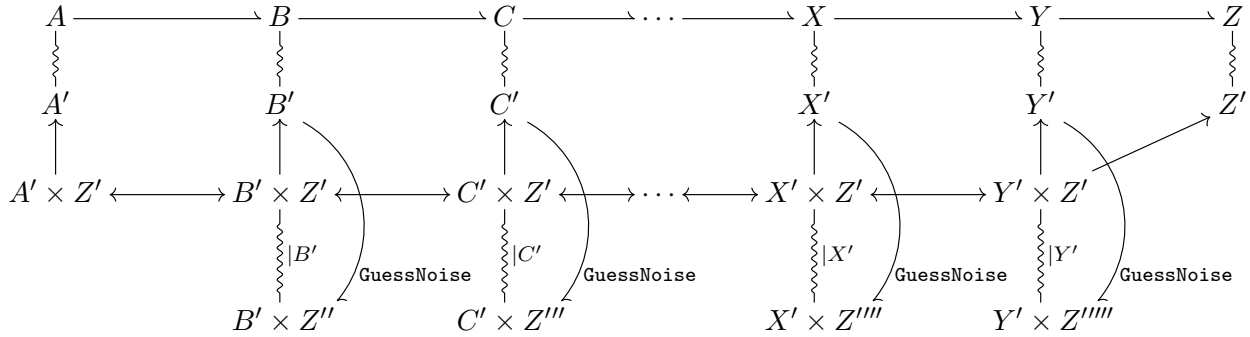
\begin{equation*}
\begin{tikzcd}[
    row sep=tiny, 
    column sep=large,
]
A \arrow[r, harpoon] & B \arrow[r, harpoon] & C \arrow[r, harpoon] &[-15pt] \cdots \arrow[r, harpoon] &[-15pt] X \arrow[r, harpoon] & Y \arrow[r, harpoon] & Z \\[2.5ex]
A' \arrow[u, no head, squigglish] & B' \arrow[u, no head, squigglish] \arrow[dd, harpoon, bend left=50, start anchor=south east, end anchor=north east, "\!\!\texttt{GuessNoise}", pos=0.8] & C' \arrow[u, no head, squigglish] \arrow[dd, harpoon, bend left=50, start anchor=south east, end anchor=north east, "\!\!\texttt{GuessNoise}", pos=0.8] & & X' \arrow[u, no head, squigglish] \arrow[dd, harpoon, bend left=50, start anchor=south east, end anchor=north east, "\!\!\texttt{GuessNoise}", pos=0.8] & Y'\arrow[u, no head, squigglish] \arrow[dd, harpoon, bend left=50, start anchor=south east, end anchor=north east, "\!\!\texttt{GuessNoise}", pos=0.8] & Z' \arrow[u, no head, squigglish] \\[2.5ex]
A' \times Z' \arrow[u] \arrow[r, <->] & B' \times Z' \arrow[u] \arrow[r, <->] & C' \times Z' \arrow[u] \arrow[r, <->] & \cdots \arrow[r, <->] & X' \times Z' \arrow[u] \arrow[r, <->] & Y' \times Z' \arrow[u] \arrow[ur] \\[5.5ex]
& B' \times Z'\mathrlap{'} \arrow[u, no head, squigglish, "|B'"'] & C' \times Z'\mathrlap{''} \arrow[u, no head, squigglish, "|C'"'] & & X' \times Z'\mathrlap{'''} \arrow[u, no head, squigglish, "|X'"'] & Y' \times Z'\mathrlap{''''} \arrow[u, no head, squigglish, "|Y'"']
\end{tikzcd}
\end{equation*}}%

Projecting $B' \times Z'$ to $B'$ destroys information about $Z'$ and, without it, we cannot get back to $A'$: this projection destroys information about $A'$.
We may thus train a \texttt{GuessNoise} model to use $B'$ to predict the ``missing noise'' $Z''$, with the intent to obtain a $B' \times Z''$ whose distribution matches $B' \times Z'$.
Equation (14) from DDPM~\autocite{ho2020denoising} learns such denoising functions.

If $\alpha_Y$ is sufficiently close to zero, we may approximately sample $Y'$ by using instead the known distribution for $Z$, then use \texttt{GuessNoise} to predict $Y'\times Z'''''$.
Now, if that $Y'\times Z'''''$ were ``perfect'', we could move horizontally all the way to $A'\times Z'''''$ by doing some algebra with $Y' = \sqrt{\alpha_Y} A'  + \sqrt{1 - \alpha_Y} Z'$.
However, approximations were involved, and we would amplify them by dividing by $\alpha_Y$ (which we just required to be close to zero).
So we do this more gradually: shift horizontally to $X'\times Z'''''$, project up to $X'$, use \texttt{GuessNoise} to predict $X'\times Z''''$, shift horizontally to $W'\times Z''''$, \emph{etc.} up to Alice.\footnote{Of course, one may also skip some steps and trade quality for speed\ldots}

Some readers may have seen multiple expositions and/or tutorials as to what DDPM is doing, and I bet that none of them looked like the above.
There are some quite subtle things happening here, and they are difficult to convey accurately through text.
I believe that diagrams such as the above enable accurate, concise, and approachable expositions.\footnote{For similar reasons, this blogpost purposefully avoids terms such as ``forward process'', ``reverse process'', or other time-related analogies, which usually abound in diffusion works. I find data processing stories about destruction and generation of information to be easier to follow than time-travel ones.}
What do you think?

\subsection{About the Diagrams}

There exist many types of \href{https://en.wikipedia.org/wiki/Graphical_model}{probabilistic graphical models}.
Some are meant to be general-purpose, others are more specialized.
Most are meant for empirical/scientific use: we translate our intuition of what are the non-impossible dependencies between stochastic variables, then we may analyze empirical observations under those assumptions.

This blogpost introduces a new kind of probabilistic graphical model, which I propose to call \emph{generative commutative diagrams}.
Unlike many other kinds of graphical models, those used here are meant for prescription/engineering use.
They read like code: my arrows and harpoons are like procedures you can call to transform the origin node into the target node.

As the name ``generative commutative diagrams'' implies, they have been inspired by the \href{https://en.wikipedia.org/wiki/Commutative_diagram}{commutative diagrams} used by mathematicians, especially \href{https://en.wikipedia.org/wiki/Category_theory}{category theorists}.
This blogpost is not about category theory, you do not need to know nor care about category theory to read it, and I am myself definitely \emph{not} an expert on category theory.

An introduction to category theory will often start with a \emph{diagram} like this.

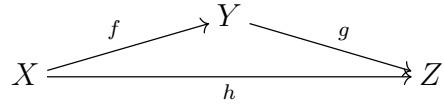
\begin{equation*}
\begin{tikzcd}[
    row sep=tiny, 
    column sep=huge,
]
    & Y \arrow[dr, "g"]\\
    X \arrow[ur, "f"] \arrow[rr, "h"'] & & Z
\end{tikzcd}
\end{equation*}

The arrows ($\rightarrow$) of such a diagram represent morphisms, which for our present purpose may very well be understood as ``deterministic function'', and this blogpost abides by this convention: whenever I use $\rightarrow$, I mean a deterministic (mathematical) function.
However, note that my generative diagrams also allow for harpoons ($\rightharpoonup$), which need not be deterministic: they may represent something less-than-a-morphism.

That same introduction to category theory will then likely say something along the lines ``The above diagram is commutative, which means that $g(f(x)) = h(x)$ for all $x$.''
When we show the ``maps to'' arrows ($\mapsto$) between the \cancel{messages} objects, this means that the paths going from $X$ to $Z$ by following $f$ then $g$ must all agree with the paths directly using $h$.

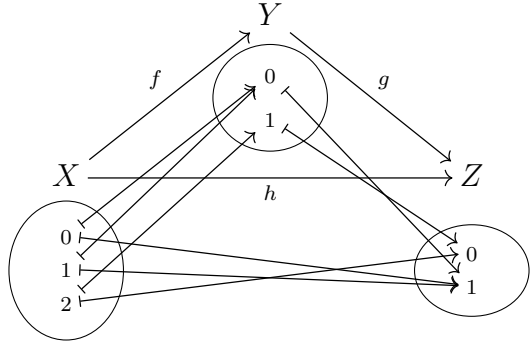
\begin{equation*}
\begin{tikzcd}[
    row sep=tiny, 
    column sep=huge,
    execute at end picture={
        \node[setellipse, fit={(\tikzcdmatrixname-5-1) (\tikzcdmatrixname-9-1)}]{};
        \node[setellipse, fit={(\tikzcdmatrixname-2-2) (\tikzcdmatrixname-3-2)}]{};
        \node[setellipse, fit={(\tikzcdmatrixname-6-3) (\tikzcdmatrixname-8-3)}]{};
    }
]
& Y \arrow[dddr, "g"] \\[.5ex]
& {\scriptstyle0} \arrow[ddddddr, mapsto] \\[-.5ex]
& {\scriptstyle1} \arrow[dddr, mapsto] \\
X \arrow[uuur, "f"] \arrow[rr, "h"'] & & Z \\[.5ex]
{\scriptstyle0} \arrow[uuur, mapsto] \arrow[dddrr, mapsto] \\[-2.5ex]
&& {\scriptstyle0} \\[-2.5ex]
{\scriptstyle1} \arrow[uuuuur, mapsto] \arrow[drr, mapsto] \\[-2.5ex]
&& {\scriptstyle1} \\[-2.5ex]
{\scriptstyle2} \arrow[uuuuuur, mapsto] \arrow[uuurr, mapsto]
\end{tikzcd}
\end{equation*}%

I define generative commutative diagrams such that both the arrows ($\rightarrow$) and harpoons ($\rightharpoonup$) \textbf{must also satisfy this requirement of commutativity.}

Next, that introduction to category theory may give the following diagram.

\begin{equation*}
\begin{tikzcd}[
    row sep=tiny, 
    column sep=huge,
]
X \arrow[r, "f"] & Y \arrow[r, "g"] & Z
\end{tikzcd}
\end{equation*}

They will then say that because $f$ and $g$ are morphisms, this picture is equivalent to the first one: we can always combine morphisms head-to-tail to obtain an implicit morphism, here $h = g \circ f$.
\textbf{This is not true for my harpoons.}
If either or both $f$ and $g$ are harpoons, the best you can get by chaining them is a way to sample the same probability distribution -- \emph{not} a guarantee that you will end up at the same element.

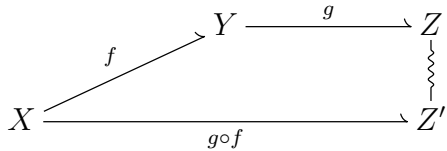
\begin{equation*}
\begin{tikzcd}[
    row sep=tiny, 
    column sep=huge,
]
    & Y \arrow[r, harpoon, "g"] & Z \arrow[d, no head, squigglish] \\[2.5ex]
    X \arrow[ur, harpoon, "f"] \arrow[rr, harpoon, "g \circ f"'] & & Z'
\end{tikzcd}
\end{equation*}

Morphisms cannot create information, only destroy or preserve it.
Harpoons can create, preserve and destroy information.

Commutative diagrams that only contain morphisms may only tell ``the destruction story'', the traditional direction of the data processing inequality.
Harpoons allow you to also tell its dual, ``the generation story''.

To be clear, we didn't get any real new narrative power: you can tell ``the generation story'' solely in terms of destruction.\footnote{Though you may have to narrate it from the perspective of Laplace's demon, a narrative device that already knows everything there is to be known.}
In that view, harpoons are just ``syntactic sugar'' for a more involved combination of morphisms.

I have no idea if there is interesting maths to be done with such harpoons and, if yes, perhaps it has already been done under some name that evaded my search.
However, as a specialized kind of graphical model, I've found them particularly useful when thinking about diffusion models and their weird edge cases.

\addcontentsline{toc}{section}{References}
\printbibliography

\addcontentsline{toc}{section}{Citing this work}
\section*{Citing this work}

For attribution in academic contexts, please cite this work as
{\small
\begin{verbatim}
Noël, Piere-André. "Destruction is a General Strategy to Learn Generation;
Diffusion's Strength is to Take it Seriously; Exploration is the Future",
ICLR Blogposts, 2026.
\end{verbatim}}

\noindent For \BibTeX{}, please use
{\small
\begin{verbatim}
@inproceedings{noel2026destruction,
  author = {No\"el, Pierre-Andr\'e},
  title = {Destruction is a General Strategy to Learn Generation;
           Diffusion's Strength is to Take it Seriously; Exploration is the Future},
  booktitle = {ICLR Blogposts 2026},
  year = {2026},
  date = {April 27, 2026},
  note = {https://iclr-blogposts.github.io/2026/blog/2026/destruction/},
  url  = {https://iclr-blogposts.github.io/2026/blog/2026/destruction/}
}
\end{verbatim}}

\end{document}